\pdfoutput=1

\documentclass[11pt,a4paper]{article}

\usepackage[acceptedWithA]{tacl2021v1}

\usepackage{times}
\usepackage{latexsym}

\usepackage[T1]{fontenc}
\usepackage{url}

\usepackage[T1]{fontenc}

\usepackage[utf8]{inputenc}

\usepackage{microtype}

\usepackage{inconsolata}

\usepackage{array}
\usepackage{amsmath, bm, amssymb, enumitem}
\usepackage{graphicx}

\title{Characterizing Learning Curves During Language Model Pre-Training: Learning, Forgetting, and Stability}

\author{Tyler A. Chang$^{1,2}$, \quad Zhuowen Tu$^1$, \quad Benjamin K. Bergen$^1$ \\
$^1$Department of Cognitive Science \\
$^2${Halıcıoğlu} Data Science Institute \\
University of California San Diego \\
{\texttt{$\{$tachang, ztu, bkbergen$\}$@ucsd.edu}}
}

\begin{document}
\maketitle
\begin{abstract}
How do language models learn to make predictions during pre-training?
To study this, we extract learning curves from five autoregressive English language model pre-training runs, for 1M unseen tokens in context.
We observe that the language models generate short repetitive phrases before learning to generate longer and more coherent text.
We also find that individual tokens often exhibit sudden increases or decreases in loss that are surprisingly consistent across pre-training runs.
To better understand these fluctuations, we quantify the final surprisal, within-run variability, age of acquisition, forgettability, and cross-run variability of learning curves for individual tokens in context.
More frequent tokens reach lower final surprisals, exhibit less variability within and across pre-training runs, are learned earlier, and are less likely to be ``forgotten'' during pre-training.
Higher $n$-gram probabilities further accentuate these effects.
Independent of the target token, shorter and more frequent contexts correlate with marginally more stable and quickly acquired predictions.
Based on our results, we argue for the existence of sequential learning dependencies between different model capabilities, and we characterize language model learning as early $n$-gram learning before gradual refinement of tail $n$-gram predictions.
\end{abstract}

\section{Introduction}
Language models have received unprecedented attention in recent years due to impressive performance on natural language tasks (e.g. \citealp{openai-2022-chatgpt,google-2023-palm2,anthropic-2023-claude}).
However, these models are initialized as random word (token) generators, and it remains unclear how the models achieve complex linguistic abilities during pre-training.
Previous work has investigated when syntactic, semantic, and reasoning abilities emerge \citep{liu-etal-2021-probing-across,evanson-etal-2023-language}, quantified ages of acquisition for tokens averaged over contexts \citep{chang-bergen-2022-word}, and extracted learning curves for individual examples \citep{xia-etal-2023-training}.
However, features that influence individual learning curves have yet to be identified (e.g. $n$-gram probabilities and context lengths).
Given any token in context, it is largely unknown when or how stably that token would be learned.

From a scientific perspective, understanding when examples are learned by language models can provide insights into possible mechanisms for language acquisition.
Regardless of their similarity to human language processing, language models are exemplars of how learning from language statistics alone (i.e. ``distributional'' learning) can lead to complex linguistic abilities \citep{chang-bergen-2022-word,warstadt-bowman-2023-artificial,mahowald-etal-2023-dissociating}.
Notably, despite smoothly decreasing corpus-level loss and independent and identically distributed (i.i.d.) data throughout pre-training, individual text examples exhibit learning curves with sudden decreases and increases in loss (\S\ref{sec:metrics} and \citealp{xia-etal-2023-training}).
This highlights the importance of examining individual example learning curves for pre-training dynamics research; aggregate curves often do not capture the fluctuations exhibited by individual examples.
Our work seeks to characterize these fine-grained convergence patterns in terms of simpler distributional statistics.

From a practical perspective, understanding language model learning curves can inform the pre-training and deployment of language models.
Learning curve results might allow NLP practitioners to determine how much pre-training is necessary for different capabilities and what behaviors will remain stable after additional pre-training (e.g. ``continual learning'' on more recent data; \citealp{jin-etal-2022-lifelong-pretraining}).
Learning curve results can also help identify scenarios in which to expect high levels of variability among fully-trained models, or even develop better pre-training curricula.
For example, better curricula might maximize the presence of tractable features that a language model can learn at different pre-training steps.

Thus, our work seeks to quantify convergence patterns for individual tokens in context during language model pre-training.
We focus on learning curve convergence, including learning speed, forgetting, and stability.
Rather than evaluate model performance on downstream tasks throughout pre-training, we study individual tokens in context (c.f. \citealp{liu-etal-2021-probing-across,xia-etal-2023-training}). 
Specifically, we run five English language model pre-training runs, and we extract learning curves for 1M unseen tokens in context.
We quantify the final surprisal, variability within and across pre-training runs, age of acquisition, and forgettability of each example.
We report general learning curve patterns, and we assess the impact of token frequencies, $n$-gram probabilities, context lengths and likelihoods, and part-of-speech tags on the speed and stability of language model learning.
Based on our results, we argue that there exist sequential dependencies between when language models acquire different capabilities (\S\ref{sec:discussion}).
We then characterize language model learning as early $n$-gram learning, before gradual refinement of low probability $n$-gram predictions based on longer context and more nuanced linguistic capabilities.
Finally, we discuss implications of our work for informed language model deployment.

\section{Related Work}
\label{sec:related-work}
Previous work has studied the pre-training dynamics of language models \citep{saphra-lopez-2019-understanding}.
\citet{choshen-etal-2022-grammar} and \citet{evanson-etal-2023-language} find that language models learn linguistic generalizations in similar stages regardless of model architecture, initialization, and data-shuffling.
In masked language models, syntactic rules are learned early, but world knowledge and reasoning are learned later and less stably \citep{chiang-etal-2020-pretrained,liu-etal-2021-probing-across}.
\citet{olsson-etal-2022-in} find that copy mechanisms (``induction heads'' for in-context learning) appear at an inflection point during pre-training.
These results establish when a variety of abilities emerge in language models.
Our work studies more fine-grained learning trajectories by evaluating individual tokens in context.

Indeed, previous work has studied how individual tokens are learned during pre-training. 
For example, word learning is highly dependent on word frequency \citep{chang-bergen-2022-word}.
Larger models memorize more examples during pre-training without overfitting \citep{tirumala-etal-2022-memorization}, but the time step that a model sees an example does not affect memorization \citep{biderman-etal-2023-pythia}.
Most similar to our work, \citet{xia-etal-2023-training} collect learning curves for individual tokens in context, finding that some examples exhibit a ``double-descent'' trend where they first increase then decrease in surprisal.
All of the studies above collect language model learning curves during pre-training, either for individual examples or targeted benchmark performance.
Here, we introduce metrics to characterize such curves, we identify general learning patterns, and we isolate text features that are predictive of learning speed and stability.

\section{Language Model Learning Curves}
We extract learning curves for 1M unseen tokens in context from five English language model pre-training runs.
Similar learning curves are computed in \citet{xia-etal-2023-training}; we extend their work by defining metrics to characterize such learning curves (\S\ref{sec:metrics}), and we identify text features that predict each metric (\S\ref{sec:predicting-metrics}).
In this way, we aim to demonstrate the connection between simple distributional statistics (e.g. $n$-gram probabilities) and language model learning.\footnote{Code is available at \url{https://github.com/tylerachang/lm-learning-curves}.}

\subsection{Models and Dataset}
\label{sec:models}
We run five autoregressive Transformer language model pre-training runs from scratch,
following the GPT-2 architecture with 124M parameters \citep{radford-etal-2019-language}.
We run five pre-training runs in order to quantify variability in learning curves across runs (\S\ref{sec:cross-run-patterns}).
For all runs, we use the same SentencePiece tokenizer trained on 10M lines of our pre-training dataset with vocabulary size 50K.

\paragraph{Dataset and training.}
We retrieve the first 128M lines of the deduplicated OSCAR English corpus \citep{abadji-etal-2021-ungoliant}.
We tokenize the corpus, concatenating lines until each sequence has length 128.
We sample $80\%$ of the resulting dataset as our pre-training dataset (5.1B tokens), leaving the remainder for evaluation and testing.
Models are trained for 1M steps with batch size 256 \citep{devlin-etal-2019-bert,chang-bergen-2022-word}.
Each model is initialized with a different random seed and uses a different shuffle of the pre-training dataset.
Pre-training details and hyperparameters are in \S\ref{app:pretraining-details}

\setlength{\belowcaptionskip}{-0.3cm}
\begin{figure}
    \centering
    \includegraphics[width=6.8cm]{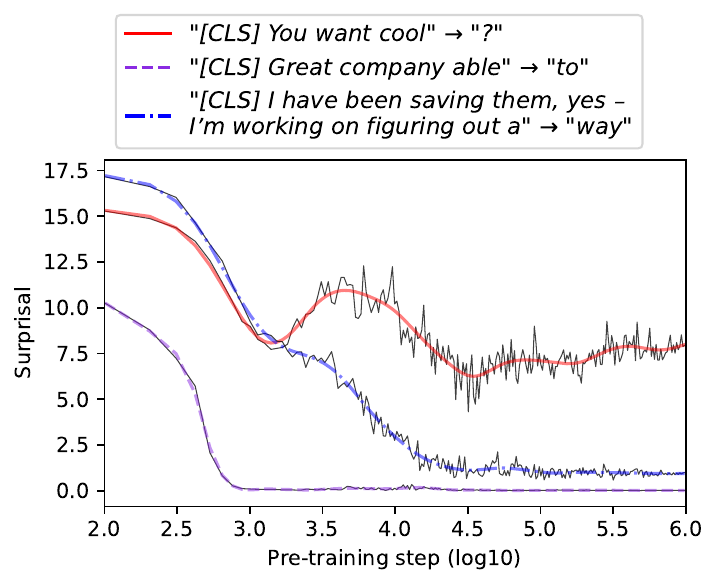}
    \vspace{-0.25cm}
    \caption{Learning curves for three evaluation examples from the OSCAR dataset during one pre-training run. Colored lines are fitted GAM curves.}
    \label{fig:surprisal-curves}
\end{figure}
\setlength{\belowcaptionskip}{0cm}

\setlength{\belowcaptionskip}{-0.2cm}
\begin{table*}[t]
    \centering
    \footnotesize
    \definecolor{purple}{HTML}{5500EE}
    \renewcommand{\arraystretch}{1.3}
    \begin{tabular}{| >\raggedleft p{0.8cm}| >\raggedleft p{1.1cm}| >{\raggedright\arraybackslash} p{12.0cm}|}
        \hline
        Step & Training tokens & Model output \\
        \hline
        0 & 0 & \texttt{``This is\color{purple}469 gush liqueur Defense trophies Jakarta Sale Berlin deservingException validate jalapeno\color{black}...''} \\
        100 & 3.3M & \texttt{``This is\color{purple},,,,,,,,,,,,,,,,,,,,,,,,,,,,,,,, the the the the,,,,,,,......\color{black}''} \\
        1K & 33M & \texttt{``This is \color{purple} a few of the first of the same of the world’s the most of the first of the the same of the first of the world.\color{black}''} \\
        10K & 330M & \texttt{``This is \color{purple} a great way to make a difference in your life.\color{black}''} \\
        100K & 3.3B & \texttt{``This is \color{purple} a very important part of the process of getting your business off the ground.\color{black}''} \\
        1M & 33B & \texttt{``This is \color{purple} a great opportunity to own a beautiful home in the desirable area of North Vancouver.\color{black}''} \\
        \hline
    \end{tabular}
    \normalsize
    \caption{Sample model outputs completing the prompt \texttt{``This is...''} at different pre-training checkpoint steps, using sampling temperature $0.3$. We also report the total number of tokens observed up to a given step; one epoch of our pre-training dataset is 5.1B tokens.}
    \label{tab:sample-outputs}
\end{table*}
\setlength{\belowcaptionskip}{0cm}

\paragraph{Checkpoints.}
Previous work studying language models during pre-training has saved model checkpoints at inconsistent intervals (e.g. every 100 steps or every power of two up to step 1000, then every 1000 steps up to step 100K, etc; \citealp{blevins-etal-2022-analyzing,chang-bergen-2022-word,sellam-etal-2022-multiberts,biderman-etal-2023-pythia,xia-etal-2023-training}).
To obtain smoother changes between checkpoints, we save checkpoints such that the number of steps between checkpoints increases linearly as a function of the current step $t$.
As a result, (1) we can define the checkpoint frequencies at the start and end of pre-training, (2) the checkpoint step is an exponential function of the checkpoint number, and (3) the number of steps per checkpoint is an exponential function of the checkpoint number.
Checkpoint strategy details are in \S\ref{app:checkpoint-strategy}.
We begin pre-training with 100 steps per checkpoint, and we end pre-training with 25K steps per checkpoint (ending at step 1M).
Including a checkpoint at step zero, this results in 222 checkpoints per pre-training run.
Sample outputs from different checkpoints are included in \S\ref{sec:overall-patterns}.

\subsection{Surprisal Curves}
\label{sec:surprisal-curves}
For quantitative analyses of language model learning curves, we sample 100K sequences from the evaluation dataset in \S\ref{sec:models}.
We sample ten tokens per sequence, and we compute the surprisal $-\text{log}_2(P(w))$ for each token $w$ based on its preceding context \citep{levy-2008-expectation}, using each language model checkpoint.
Surprisal is an established information-theoretic metric used to measure the ``surprise'' of a next token given a language model \citep{levy-2008-expectation,goodkind-bicknell-2018-predictive,futrell-etal-2019-neural,li-etal-2021-bert,chang-bergen-2022-word,oh-schuler-2023-surprisal,michaelov-etal-2024-strong}.
We then have a learning curve for each token in context (i.e. each example) and each model, usually trending from higher surprisal (worse predictions) to lower surprisal (better predictions; Figure \ref{fig:surprisal-curves}).
Surprisal is equivalent to the language modeling loss function in log base two.
In total, we collect surprisal curves for 1M examples per model.

\section{Overall Learning Patterns}
\label{sec:overall-patterns}
Before considering fine-grained learning patterns for individual surprisal curves, we observe several overall trends during language model pre-training.
Many of these trends echo results from previous work (e.g. $n$-gram learning in \citealp{chang-bergen-2022-word,choshen-etal-2022-grammar}) or intuitive results known by language model pre-training practitioners (e.g. the slow development of the ability to generate long coherent text), but these trends establish basic intuitions about how language models progress throughout pre-training.

\begin{figure}
    \centering
    \includegraphics[width=7.0cm]{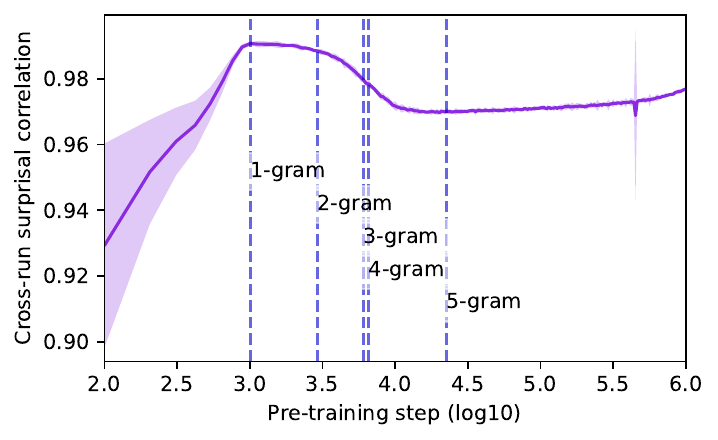}
    \vspace{-0.25cm}
    \caption{Mean pairwise correlation between model surprisals for different pre-training runs, at different pre-training steps.\protect\footnotemark~Shaded regions indicate five standard deviations from the mean.
    Vertical lines indicate the pre-training steps where model surprisals are maximally correlated with $n$-gram surprisals.}
    \label{fig:surprisal-correlations}
\end{figure}

\paragraph{Early in pre-training, models generate short repetitive phrases.}
Sample outputs from different model checkpoints are shown in Table \ref{tab:sample-outputs}.
We manually inspect outputs from all five pre-training runs, generating text completions to 100 randomly sampled subsequences from the evaluation dataset in \S\ref{sec:models}, using sampling temperature $0.3$ \citep{holtzman-etal-2019-curious}.
As expected, models initialize with random token predictions at step zero.
By 100 steps, they repeatedly produce frequent tokens; at this stage, $99.8\%$ of output tokens are ``\textit{the}'', a comma, or a period.
The remaining tokens are frequent words such as ``\textit{to}'', ``\textit{of}'', and ``\textit{and}''.
By 1000 steps, the models repeatedly produce frequent short phrases such as ``\textit{of the first}'' or ``\textit{and the most}''; $86.5\%$ of completions contain the phrase ``\textit{of the first}'', and $71.1\%$ of completions include it at least twice.
These observations align with previous work finding that language models overfit to unigram then bigram next-token predictions early in pre-training (\citealp{chang-bergen-2022-word}; see also Figure \ref{fig:surprisal-correlations}); here, we demonstrate these findings in longer sequences of generated text.

\paragraph{Models later generate longer and more coherent text.}
By step 10K, the models generally produce coherent sentence completions, but they still contain repetitive phrases ($10.8\%$ of completions with a three-word phrase repeated at least three times).
By step 100K, the repetition rate drops to $6.0\%$, and completions appear more specific to the context.
By step 1M, the repetition rate is $4.7\%$, and the models can produce coherent multi-sentence completions.
Still, due to our relatively small model size (124M parameters, the size of the original GPT model; \citealp{radford-etal-2018-improving}), we do not expect our models to exhibit text generation capabilities at the level of larger language models.

\paragraph{Models roughly follow \textit{n}-gram learning.}
We compute the correlation between $n$-gram surprisals and model surprisals throughout pre-training.\footnote{We compute $n$-gram probabilities directly from the pre-training dataset, as in \S\ref{sec:predictors-regressions}. For unobserved $n$-grams, we use backoff to $(n-1)$-grams \citep{katz-1987-estimation}.}
Consistent with previous work (\citealp{chang-bergen-2022-word}; \citealp{karpathy-etal-2015-visualizing} for LSTMs), the models overfit to unigram (token frequency) predictions then bigram predictions early in pre-training.
Extending this up to 5-grams, the models reach maximal similarity to a unigram model around step 1K, before peaking in similarity to 2, 3, 4, and 5-grams, in that order (Figure \ref{fig:surprisal-correlations}).
This is consistent with the hypothesis that language models at some specified level of performance make similar generalizations regardless of architecture \citep{choshen-etal-2022-grammar,xia-etal-2023-training}.
Figure \ref{fig:surprisal-correlations} demonstrates that as the models pre-train, their individual predictions pass through stages where they loosely match different $n$-gram models.

\paragraph{Models are maximally similar early and late in pre-training.}
We also compute the correlation between model surprisals across pre-training runs at different checkpoints (Figure \ref{fig:surprisal-correlations}).
At any given checkpoint, the similarity between any two pre-training runs is both high (Pearson's $r>0.95$ after step 1K) and consistent (extremely low standard deviations; Figure \ref{fig:surprisal-correlations}).
The models are maximally similar almost exactly when they mirror the unigram distribution (i.e. predicting based on token frequency).
The models then decrease slowly in cross-run similarity, reaching a local minimum as they approach the 5-gram distribution.
This suggests that there is at least some variability in the generalizations that language models make beyond bigrams.
Still, as demonstrated by the steady increase in similarity throughout the remainder of pre-training, language models eventually converge to similar solutions as their performance improves.

\setlength{\belowcaptionskip}{-0.3cm}
\setlength{\belowcaptionskip}{0cm}
\footnotetext{At approximately $10^{5.7}$ steps, one model exhibited a small temporary increase in loss, leading to a dip in the cross-run surprisal correlation.}

\section{Characterizing Learning Curves}
\label{sec:metrics}
We then consider fine-grained analyses of learning curves for individual tokens in context.
We introduce five metrics to characterize language model learning curves, each motivated by previous work.

\subsection{Within-Run Metrics}
\label{sec:within-run-metrics}
First, we compute four metrics for each learning curve within a pre-training run (\S\ref{sec:surprisal-curves}): final surprisal, variability across pre-training steps, age of acquisition, and forgettability.

\paragraph{Final surprisal.}
Surprisal quantifies the quality of a language model's predictions for a token in context, with lower values corresponding to better predictions (\citealp{levy-2008-expectation}; \S\ref{sec:surprisal-curves}).
For each example, we compute the mean surprisal during the last 25\% of pre-training.
This is closely (and inversely) related to model confidence, which \citet{swayamdipta-etal-2020-dataset} define as the mean probability assigned to the correct label for an example during language model fine-tuning.
We use surprisals (i.e. negative log probabilities) instead of raw probabilities because the language modeling task has a much larger number of output labels (50K possible next tokens) than traditional classification tasks, leading to much lower output probabilities.
Surprisal enables distinctions amongst lower probabilities, and it is commonly used for language modeling (\S\ref{sec:surprisal-curves}).

\paragraph{Variability (steps).}
We then measure how much model performance for an example changes across steps within a pre-training run.
Specifically, we consider variability late in pre-training, when a language model has largely converged.
Longer term fluctuations in performance are captured by forgettability, defined later in this section.
Motivated by \citet{swayamdipta-etal-2020-dataset}, who compute the standard deviation of model probabilities during fine-tuning, we compute the standard deviation of surprisal during the last 25\% of pre-training.

\begin{figure}
    \centering
    \includegraphics[width=6.8cm]{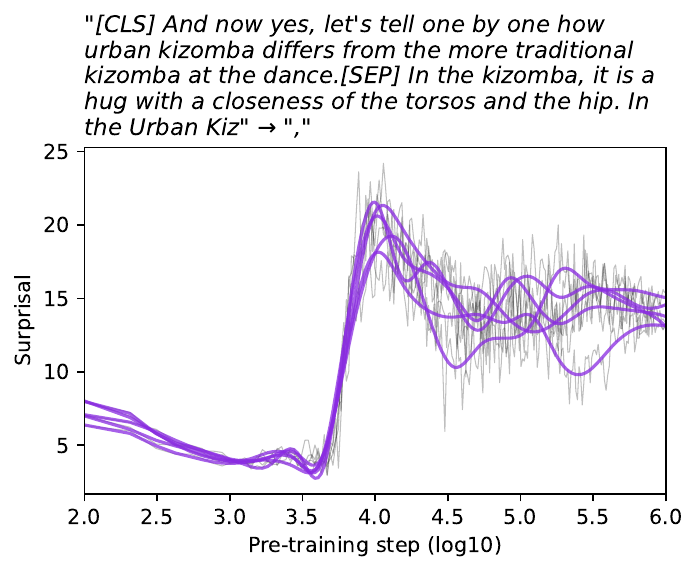}
    \vspace{-0.25cm}
    \includegraphics[width=6.8cm]{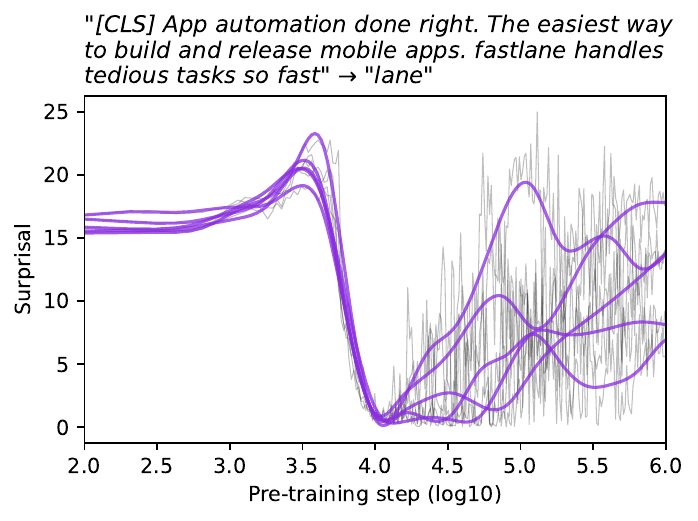}
    \caption{Learning curves for two evaluation examples from the OSCAR dataset with high forgettability scores, for the five pre-training runs. Purple lines are fitted GAM curves, one per pre-training run.}
    \label{fig:forgettability_curves}
\end{figure}

\paragraph{Age of acquisition (AoA).}
We also measure when each example is learned during pre-training.
\citet{chang-bergen-2022-word} define a token's age of acquisition (AoA) in a language model as the log-pre-training step when the model's surprisal reaches 50\% between random chance surprisal and the minimum surprisal attained by the model.
\citet{chang-bergen-2022-word} fit a sigmoid curve to the mean surprisal curve over all occurrences of the token.
Because surprisal curves for individual examples are less stable than mean curves (e.g. sometimes exhibiting both peaks and dips in surprisal; Figures \ref{fig:surprisal-curves} and \ref{fig:forgettability_curves}), we instead fit a GAM curve to each surprisal curve (surprisal $\sim$ log-pre-training step).\footnote{We fit linear GAMs with 25 splines. These are smoothed piecewise functions with 25 linear segments \citep{wood-2017-gams,pygam-2018}.}
We define an example's age of acquisition as the log-pre-training step where the fitted GAM first passes 50\% between random chance surprisal and the GAM's minimum surprisal.

\setlength{\belowcaptionskip}{-0.3cm}
\setlength{\belowcaptionskip}{0cm}

\paragraph{Forgettability.}
Along with short-term surprisal spikes as quantified by variability (across steps), language models exhibit long-term increases in surprisal for some examples during pre-training \citep{xia-etal-2023-training}.
This process is described as ``forgetting''.
To quantify long-term surprisal increases, we measure the total surprisal increase along the GAM curve fitted to each surprisal curve.
Equivalently, this is the total surprisal difference between each relative maximum and its preceding relative minimum in the curve.
Larger values indicate that an example is ``forgotten'' to a larger extent at some point during pre-training.
Example curves with high forgettability scores are shown in Figure \ref{fig:forgettability_curves}.

\subsection{Across-Run Metrics}
\label{sec:cross-run-patterns}
\paragraph{Individual learning curves are similar across pre-training runs.}
Each of the metrics in \S\ref{sec:within-run-metrics} correlates across pre-training runs ($r=0.652$ to $0.978$; diagonal entries in Table \ref{tab:correlations}).
Curves for a given example even exhibit similar peaks and dips across pre-training runs (Figure \ref{fig:forgettability_curves}).
Concretely, we quantify the distance between learning curves for two pre-training runs using the Euclidean distance between their fitted GAM curves.
Given an example curve in one pre-training run, the curve for the same example in another pre-training run is on average (median) closer than the curve for 99.93\% of other examples.\footnote{We obtain similar results using distances between raw surprisal curves. Raw surprisal curve distances are highly correlated with fitted GAM curve distances ($r=0.964$).}

\paragraph{Variability (runs).}
However, learning curves are not identical across runs.
To quantify the cross-run variability of learning curves for a given example, we compute the mean pairwise distance (squared Euclidean distance) between the fitted GAM curves for different pre-training runs.
This metric is correlated when computed using different three-run subsets of the five pre-training runs ($r=0.798$; Table \ref{tab:correlations}).
Our final cross-run variability metric is computed over all five pre-training runs.

\setlength{\belowcaptionskip}{-0.3cm}
\begin{table}[t]
    \centering
    \footnotesize
    \renewcommand{\arraystretch}{1.3}
    \begin{tabular}{|p{2.2cm}|ccccc|}
        \cline{2-6}
        \multicolumn{1}{c|}{} &
        \multicolumn{1}{c|}{\rotatebox{90}{ Surprisal }} &
        \multicolumn{1}{c|}{\rotatebox{90}{ Var. (steps) }} &
        \multicolumn{1}{c|}{\rotatebox{90}{ AoA }} &
        \multicolumn{1}{c|}{\rotatebox{90}{ Forgettability\phantom{\_} }} &
        \multicolumn{1}{c|}{\rotatebox{90}{ Var. (runs) }} \\ 
        \hline
        Surprisal & \textbf{0.98} & 0.46 & 0.31 & 0.62 & 0.45 \\
        Variability (steps) & & \textbf{0.65} & 0.38 & 0.43 & 0.57 \\
        AoA & & & \textbf{0.84} & 0.14 & 0.43 \\
        Forgettability & & & & \textbf{0.79} & 0.51 \\
        Variability (runs) & & & & & \textbf{0.80} \\
        \hline
    \end{tabular}
    \normalsize
    \caption{Pearson correlations between learning curve metrics. Diagonal entries indicate the mean correlation for that metric across pre-training runs. For variability across runs, the diagonal entry is the mean correlation between cross-run variability scores computed from different three-run subsets of the five pre-training runs.}
    \label{tab:correlations}
\end{table}
\setlength{\belowcaptionskip}{0cm}

\subsection{Correlations Between Metrics}
\paragraph{Surprisal correlates with all learning curve metrics.}
Correlations between metrics are reported in Table \ref{tab:correlations}.
All five metrics are positively correlated with one another.
High-surprisal examples exhibit more variability across pre-training steps, are learned later, are more likely to be forgotten during pre-training, and exhibit more cross-run variability.
Some of these correlations are unsurprising based on our metric definitions; for example, forgettability is quantified using surprisal curve increases during pre-training, which likely lead to higher final surprisals.
However, the correlation between final surprisal and forgettability is far from perfect ($r=0.622$), suggesting that some examples can be forgotten and then re-learned (high forgettability, low surprisal) or simply never learned (low forgettability, high surprisal).
Indeed, upon manual inspection, we observe both of these types of curves.
Of the $269$ examples in both the top 5\% of forgettability and bottom 5\% of surprisal, $92\%$ exhibit a sudden (greater than $2.5$) surprisal increase in the fitted GAM curve that is later recovered. Of the $32$ examples in both the bottom 5\% of forgettability and top 5\% of surprisal, $78\%$ never deviate from their starting surprisal by more than $20\%$.

\begin{table*}[t]
    \centering
    \footnotesize
    \renewcommand{\arraystretch}{1.3}
    \begin{tabular}{|>\raggedright p{4.0cm}|
    >\raggedleft p{1.6cm}| >\raggedleft p{1.6cm}| >\raggedleft p{1.6cm}| >\raggedleft p{1.7cm}| >{\raggedleft\arraybackslash} p{1.6cm}|}
        \hline
        Predictor & Surprisal & Var. (steps) & AoA & Forgettability & Var. (runs) \\
        \hline
        Target token log-frequency & $R^2=$ 0.268 & $R^2=$ 0.248 & $R^2=$ 0.763 & $R^2=$ 0.083 & $R^2=$ 0.195 \\
        + Target 5-gram log-prob & (--) \hspace{0.1cm} + 0.325 & (--) \hspace{0.1cm} + 0.050 & (+) \hspace{0.1cm} + 0.001 & (--) \hspace{0.1cm} + 0.149 & (--) \hspace{0.1cm} + 0.042 \\
        + Context log-length & (--) \hspace{0.1cm} + 0.007 & (+) \hspace{0.1cm} + 0.005 & (+) \hspace{0.1cm} + 0.001 & (+) \hspace{0.1cm} + 0.002 & (+) \hspace{0.1cm} + 0.005 \\
        + Context 1-gram log-prob & (+) \hspace{0.1cm} + 0.001 & (--) \hspace{0.1cm} + 0.006 & (--) \hspace{0.1cm} + 0.001 & (--) \hspace{0.1cm} + 0.010 & (--) \hspace{0.1cm} + 0.012 \\
        + Target contextual diversity & (+) \hspace{0.1cm} + 0.003 & (+) \hspace{0.1cm} + 0.000 & (+) \hspace{0.1cm} + 0.000 & (+) \hspace{0.1cm} + 0.001 & (+) \hspace{0.1cm} + 0.001 \\
        + Target part-of-speech & + 0.009 & + 0.006 & + 0.014 & + 0.028 & + 0.026 \\
        \hline
        Total variance accounted & 61.2\% & 31.5\% & 78.1\% & 27.2\% & 28.1\% \\
        \hline
    \end{tabular}
    \normalsize
    \caption{Increases in adjusted $R^2$ values when predicting each learning curve metric, iteratively adding predictors to a linear regression. The $(+)$ symbol indicates a positive coefficient for that predictor, evaluated in three regressions (\S\ref{sec:predictors-regressions}). All other coefficients are negative. Coefficients for different part-of-speech tags are described in \S\ref{sec:results}. In the bottom row, we report the total variance accounted for in each learning curve metric using all six predictors.}
    \label{tab:regressions}
\end{table*}

\section{Predicting Learning Curve Metrics}
\label{sec:predicting-metrics}
In the previous section, we defined five metrics to characterize language model learning curves.
Next, we predict each metric from specific features of each example, including $n$-gram probabilities, context likelihoods, and part-of-speech tags.
We use a linear regression to quantify effects over all 1M examples, providing evidence that simple text features can predict language model learning patterns.

\subsection{Predictors and Regressions}
\label{sec:predictors-regressions}
Each text example consists of an input context and a target token (\S\ref{sec:surprisal-curves}).
We consider six predictors (text features) that may be predictive of learning curve metrics:
\begin{itemize}[leftmargin=0.4cm]
\setlength\itemsep{0.05cm}
\item Target token log-frequency: We compute the log-frequency (i.e. unigram log-probability) of the target token in the pre-training dataset.
\item Target token 5-gram log-probability: To capture the likelihood of the target token based on local context, we compute the log-probability of the target token conditioned only on the previous four tokens (i.e. a 5-gram model). We compute probabilities directly from the pre-training dataset, and we use backoff to $(n-1)$-grams when an $n$-gram is not observed in the dataset \citep{katz-1987-estimation}. Because 5-gram log-probability is roughly linearly related to target token log-frequency ($r=0.632$), we compute the 5-gram log-probability residuals after regressing over target log-frequency.
This captures the 5-gram log-probability after accounting for target token log-frequency.
\item Context log-length: We compute the log of the number of context tokens.
\item Context log-probability:
We also compute the likelihood of the context, independent of the target token.
We compute the mean log-frequency of all context tokens, equal to the negative log-perplexity of the context using a unigram language model.
We use a unigram model to capture context frequency independent of word order within the context \citep{blei-etal-2003-latent}; longer $n$-gram models are more likely to capture probabilities of specific local constructions, even when they are distant from the target token.
\item Target token contextual diversity: The diversity of contexts in which a word appears influences word learning in people, with beneficial effects in adults but potentially hindering effects in young children \citep{hills-etal-2010-associative,johns-etal-2016-influence,rosa-etal-2022-contextual,chang-bergen-2022-contextual}.
As in \citet{hills-etal-2010-associative}, we count the number of unique tokens that appear within 30 tokens of the target token in the pre-training dataset.\footnote{Because our language models are autoregressive, we only consider context tokens that appear before the target token. We restrict our counts of co-occurring tokens to the 10K most frequent tokens in the dataset \citep{hills-etal-2010-associative}.}
To remove a nonlinear effect of token frequency on this raw diversity metric, we compute the residuals after fitting a GAM curve predicting a token's contextual diversity from its log-frequency \citep{chang-bergen-2022-contextual}.
These residuals serve as a frequency-adjusted measure of a token's contextual diversity.
\item Target token part-of-speech (POS): We annotate each example with POS tags (e.g. nouns, verbs, and adjectives; \S\ref{app:pos-coefs}) using spaCy \citep{honnibal-etal-2020-spacy}, and we consider the POS tag of the target token.
Because words can span multiple tokens, we include a feature indicating whether the target token is the first token, intermediate token, last token, or only token in a word.
\end{itemize}
We fit separate linear regressions predicting each learning curve metric from the predictors above, iteratively adding predictors in the order listed.\footnote{We exclude interaction terms, which we find do not substantially improve predictions.
Adjusted $R^2$ values increase by less than $0.03$ even when including an interaction term between every pair of continuous predictors. We clip each predictor to five standard deviations from the mean.}
We fit each regression to all 1M examples, predicting the mean value of each learning curve metric over all pre-training runs.
We run likelihood ratio tests to assess whether each predictor is predictive of the target metric after accounting for all previous predictors, but we find that every test is highly significant ($p<0.0001$).
This is likely because the large number of examples (1M) makes even small effects statistically significant.
Thus, we report adjusted $R^2$ values that capture the magnitude of effect of each predictor, after accounting for previous predictors (Table \ref{tab:regressions}).

To assess the direction of effect for each continuous predictor on each learning curve metric, we consider the coefficient for that predictor in (1) a regression containing all predictors, (2) a regression containing that predictor alone, and (3) a regression containing that predictor alone but accounting for token log-frequency in the target metric (i.e. predicting learning curve metric residuals after the log-frequency regression).
In all but one case, we obtain the same direction of effect in all three regressions.\footnote{We obtain a negative coefficient for contextual diversity in one of three cases when predicting within-run variability. All other coefficients for contextual diversity are positive (\S\ref{sec:results}).}
Furthermore, the Pearson correlation between each pair of predictors is less than $r=0.2$, and the variance inflation factor (VIF) for each predictor is less than $1.1$.
This indicates that the signs of our regression coefficients are safely interpretable.
For effects of POS (a categorical variable), we consider the regression coefficient for different POS tags after accounting for all other predictors, by predicting learning curve metric residuals after regressing over the other predictors.

\subsection{Results}
\label{sec:results}
The following conclusions are based on the regression results quantifying effects over all 1M examples.
Results are reported in Table \ref{tab:regressions}, including the direction of effect for each predictor and the variance accounted for in each learning curve metric.

\paragraph{Target token log-frequency.}
Frequent target tokens reach lower surprisals, are acquired faster, exhibit less variability within and across pre-training runs, and are less likely to be forgotten during pre-training.
This is consistent with previous work showing that language models are highly reliant on token frequencies for syntactic rule learning \citep{wei-etal-2021-frequency}, numerical reasoning \citep{razeghi-etal-2022-impact}, and overall word learning \citep{chang-bergen-2022-word}.
Our work indicates that this effect persists at the individual example level.

\paragraph{Target 5-gram log-probability.}
Unsurprisingly, 5-gram log-probabilities correlate with lower final surprisals after accounting for target token frequency; in other words, predictions from a 5-gram model and a Transformer model are correlated beyond the effects of token frequency.
More notably, higher 5-gram log-probabilities are predictive of lower learning variability both within and across pre-training runs, along with lower forgettability.
The added effect of 5-gram log-probability on forgettability ($+0.149$ $R^2$) is even stronger than the effect of target token frequency alone ($0.083$ $R^2$), suggesting that conditional token probabilities play a more significant role in language model forgetting than raw token frequencies.

Less intuitively, higher 5-gram log-probabilities are correlated with marginally later ages of acquisition.
We hypothesize that this is because 5-grams do take time to learn (Figure \ref{fig:surprisal-correlations}), but low probability 5-grams are more likely to never be learned at all, reaching their minima early in training (e.g. during the unigram learning phase).
This could drive the small effect where low probability 5-grams appear to be learned earlier.
Indeed, of the $122$ examples in both the bottom 1\% of 5-gram log-probabilities and the earliest 1\% of AoAs, $89\%$ reach their minimum surprisal during the first 1K steps but then exhibit substantial (greater than $2.5$) increases and fluctuations in surprisal for the remainder of pre-training.
Notably, $96\%$ never improve from random chance surprisal by more than $5\%$.
In other words, low 5-gram probability examples may appear to exhibit early AoAs, but this is primarily because they are never learned particularly well, not due to early learning curve convergence.
This reflects the fact that surprisal curves are not always accurate measures of ``learning'' (\S\ref{sec:discussion}).
An early drop in surprisal does not always indicate that an example is ``learned''.

\paragraph{Context log-length.}
The remaining predictors account for far less variance in learning curve metrics than target log-frequency and 5-gram log-probability.
Longer contexts correlate with lower surprisals, indicating that models successfully incorporate information from preceding context.
However, longer contexts also correlate with higher variability within and across pre-training runs, higher forgettability, and later AoAs.
This may be because predictions for a highly specific context are less generalizable and are thus learned less robustly by the models.
This instability for long-context predictions is particularly notable as language models are increasingly used with long contexts (e.g. full conversations; \citealp{openai-2022-chatgpt}).

\paragraph{Context log-probability.}
More frequent contexts are predictive of lower variance within and across pre-training runs, earlier acquisition, and lower forgettability.
When models are repeatedly exposed to a context, regardless of the target token, their predictions stabilize earlier and with less variability.
However, more frequent contexts also correlate with higher surprisals, indicating overall ``worse'' predictions.
This may be because frequent contexts (e.g. descriptions of common situations) on average impose fewer constraints on the next token, leading to more ambiguous ground truth distributions and thus higher surprisals.
If this is the case, the optimal surprisal values are simply higher in frequent contexts, but the models still learn faster and more stably given these contexts.

We note that the directions of effect for context log-probability remain stable for different window sizes of preceding context.
After regressing out target token log-frequency, every coefficient sign for context log-probability remains the same for all window sizes in $\{1,2,4,...,128\}$.
However, despite these consistent effects, context log-probability accounts for less than $3\%$ of the variance in each learning curve metric in all cases, even before accounting for other predictors.
Frequent contexts consistently correlate with faster and more stable learning, but with only small effects.

\paragraph{Target contextual diversity.}
Effects of contextual diversity are extremely small but statistically significant (\S\ref{sec:predictors-regressions}).
Tokens that appear in diverse contexts have higher final surprisals, are learned later, have greater variability within and across pre-training runs, and are more likely to be forgotten.
This aligns with findings that contextual diversity hinders word learning in young children \citep{chang-bergen-2022-contextual}, contrasting with results in older children and adults \citep{johns-etal-2016-influence,rosa-etal-2022-contextual}.
Diverse contexts are thought to add noise to the early word learning process, introducing an excess of possible interpretations for a word.

\paragraph{Target part-of-speech (POS).}
After accounting for other predictors, the POS tag of the target token has a small effect on each learning curve metric.
Coefficients for all POS tags are reported in \S\ref{app:pos-coefs}.
Nouns, pronouns, and punctuation symbols reach lower final surprisals than verbs, adjectives, adverbs, and interjections.
However, nouns are learned slower and with more variability (within and across pre-training runs) than adjectives, adverbs, and verbs, and they are more likely to be forgotten.
Similarly, punctuation symbols exhibit high variability and forgettability, although they are learned early and reach low surprisals.
Despite their high surprisals, interjections are learned early and stably.
These results indicate that POS tags with lower surprisals are not necessarily learned more stably.
Additionally, we find that different types of function words (e.g. conjunctions, prepositions, and determiners) have inconsistent effects, but they overall tend to be learned with high variability and forgettability.

The position of a token within a word also impacts learning curve metrics.
Sub-word tokens after the first token in a word have low final surprisals, but they exhibit high forgettability and cross-run variability.
Single-token words are the least likely to be forgotten and have the lowest cross-run variability.
Compared to the POS tag of a word, a token's position within a word has only tiny effects on within-run variability and AoA  (judged by the $R^2$ increase from including within-word position vs. only POS tag itself).
These results underline the importance of tokenizer quality in language model pre-training \citep{rust-etal-2021-good}; sub-word tokens are more likely to exhibit unstable learning despite low surprisals.

\section{Discussion}
\label{sec:discussion}
In the previous sections, we report general patterns during language model pre-training (\S\ref{sec:overall-patterns}), define ways to characterize learning curves (\S\ref{sec:metrics}), and isolate specific features that predict the speed and stability of learning for individual tokens in context (\S\ref{sec:predicting-metrics}).
Our results contribute to ongoing work studying language model pre-training dynamics, with implications for robust model deployment.

\paragraph{Sequential learning.}
Previous work has demonstrated that language models exhibit fine-grained learning patterns that are not captured by the corpus-level loss curve (related work in \S\ref{sec:related-work}).
In particular, sudden increases and decreases in example loss (\S\ref{sec:metrics} and \citealp{xia-etal-2023-training}) may be somewhat surprising given that the pre-training text is i.i.d. for all pre-training steps.
By demonstrating that many of these sudden changes are consistent regardless of random initialization and data shuffling (\S\ref{sec:cross-run-patterns}), our work indicates that some instances of sudden learning and ``forgetting'' are not due to random chance or the specific examples observed in a given step.\footnote{In our case, ``forgetting'' does not always indicate a decrease in model quality, but rather that the model has changed its output distribution such that a given ground truth token is less likely. The model distribution might still be a better reflection of text distributions overall.}
Rather, they reflect some change in model processing that consistently occurs partially into pre-training (roughly step $t \neq 0$).
Because such a sudden change cannot be attributed to the specific examples observed (robust to random shuffling) or any change in the pre-training distribution at time $t$ (the data is always i.i.d.), the primary remaining explanation is that the models' sudden ``learning'' at step $t \neq 0$ is made possible by some systematic difference between models (and their optimizers) just before step $t$ vs. at step $0$.

Framed from a potentially more interesting perspective, some types of language model ``learning'' appear to be dependent on previous learning and the linguistic abilities already present in the model.
This aligns with previous work showing that language models acquire linguistic abilities in a systematic order during pre-training \citep{liu-etal-2021-probing-across,choshen-etal-2022-grammar}, although not necessarily due to sequential dependencies.
For example, \citet{evanson-etal-2023-language} show that despite similar acquisition orders across models, different syntactic abilities are learned in parallel; performance for most individual abilities increases from the onset of pre-training.
Our work provides evidence that there exist other capabilities or types of generalizations (e.g. non-syntactic abilities or even more fine-grained syntactic sub-abilities) that can only be learned after others, or at least only once the model reaches some particular state.
Isolating these sequential dependencies is an exciting direction for future work.

\paragraph{\textit{N}-gram learning and refinement.}
As a further step towards understanding fine-grained learning patterns in language models, our work investigates whether simple statistical regularities can explain learning patterns such as the sudden loss changes discussed above.
We demonstrate that learning curves are more stable and converge faster for frequent tokens, $n$-gram probable tokens, and frequent contexts (\S\ref{sec:results}).
High probability $n$-grams in particular are less likely to be ``forgotten'', suggesting that evolving model generalizations throughout pre-training have larger effects on low-probability $n$-grams. 
Combined with findings that language models roughly follow $n$-gram learning early in pre-training and only later produce longform coherent text (\S\ref{sec:overall-patterns}; \citealp{chang-bergen-2022-word}), language model learning might be characterized as early $n$-gram learning, then gradual refinement of the tail $n$-gram probabilities based on longer contexts and more nuanced linguistic capabilities (e.g. world knowledge and reasoning; \citealp{liu-etal-2021-probing-across}).

\paragraph{Robust model deployment.}
Our work also has implications for robust model deployment.
High token frequencies and $n$-gram probabilities are by far the most influential predictors of early and stable learning in language models (\S\ref{sec:results}, with marginal additional effects of context lengths and likelihoods).
As language models are deployed in domains with highly-specific vocabulary terms (e.g. healthcare, law, and finance; \citealp{yang-etal-2023-harnessing}), the accurate prediction of infrequent domain-specific terms during text generation is likely to require extensive pre-training (late acquisition, likely mitigated by large pre-training datasets).
Such domain-specific text generation is also likely to be unstable across models and pre-training steps (high variability, potentially more difficult to mitigate).
Even if model deployment in these areas is beyond researchers' control, realistic expectations of when models might behave unstably are important to facilitate safe use by the public.
Of course, it is also possible that fine-tuning or careful prompting may reduce instability across models and training steps in these domains.

Finally, our work demonstrates that loss curves for individual examples often fluctuate in ways that are not evident from aggregate loss curves.
Even as models appear to converge (smoothly plateauing loss), models may still be adjusting predictions for tail examples in substantial ways.
Our work provides insights and methods to identify examples that are likely to exhibit late fluctuations in language model pre-training; for example, low probability $n$-grams correlate with high variability and forgettability metrics.
When determining whether a model is sufficiently and stably trained for a given use case, convergence for these types of examples should be considered.

\paragraph{Limitations and scaling.}
Our work has several limitations.
First, surprisal is an imperfect proxy for language model learning.
A model might achieve the same surprisal at different points during pre-training by using different internal prediction strategies (e.g. predicting the same token based on frequency vs. more nuanced reasoning).
Additionally, reaching some minimum surprisal does not mean that an example is ``learned''; it simply indicates the best performance achieved by a model.
The optimal surprisal is not necessarily zero due to the nondeterminism of language.
That said, surprisal remains a common measure of language model behavior \citep{futrell-etal-2019-neural,li-etal-2021-bert}, performance \citep{hoffmann-etal-2022-training}, and learning \citep{chang-bergen-2022-word,xia-etal-2023-training}, and it requires no annotated text data to compute.

Second, we only consider language models with 124M parameters trained on 5.1B tokens.
Previous work has demonstrated that learning curves differ across model sizes \citep{xia-etal-2023-training}; larger models are able to ``learn'' some examples (usually late in pre-training) for which smaller models reach non-optimal local minima or even diverge.
Larger models also exhibit less forgetting of pre-training examples \citep{tirumala-etal-2022-memorization}, although it remains unclear whether similar mechanisms are responsible for evaluation example forgetting (i.e. surprisal increases for seen vs. unseen examples).
Further research is necessary to determine the effects of model size on learning speed, variability, and forgetting; with a larger compute budget, the methods presented in our work can easily be applied to larger models.
Nonetheless, previous work has documented similar behaviors for different model sizes when they achieve similar perplexities \citep{choshen-etal-2022-grammar,xia-etal-2023-training}, suggesting that pre-training dynamics in smaller models may be similar to the early dynamics of larger models.
A particularly exciting direction for future work is to characterize the examples (e.g. based on types of reasoning, world knowledge, or commonsense) that fluctuate at different points during pre-training across model sizes.

\section{Conclusion}
In this work, we identify learning patterns during language model pre-training, including concrete features that predict when and how stably individual examples are acquired.
We assess the impact of $n$-gram probabilities, context lengths and likelihoods, and part-of-speech tags on the speed and stability of language model learning.
We propose a high-level characterization of language model learning based on simple distributional statistics, and we discuss implications for deploying robust language models in practice.

\section*{Acknowledgements}
We thank the UCSD Language and Cognition Lab for valuable discussion, and the anonymous reviewers for insightful comments.
Some models were trained on hardware provided by the NVIDIA Corporation as part of an NVIDIA Academic Hardware Grant.
Zhuowen Tu is supported by NSF IIS-2127544.
Tyler Chang is partially supported by the UCSD HDSI graduate fellowship.

\bibliography{anthology,custom}
\bibliographystyle{acl_natbib}

\appendix

\section{Appendix}
\label{app:appendix}

\subsection{Pre-Training Details}
\label{app:pretraining-details}
\begin{figure}[ht]
    \centering
    \includegraphics[width=5.8cm]{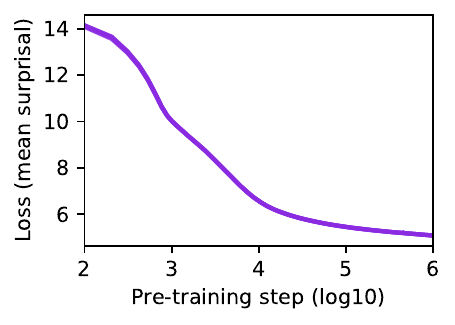}
    \vspace{-0.25cm}
    \caption{Loss curves (mean surprisal) for all five pre-training runs. Loss curves are nearly identical across runs. To align with other figures, pre-training steps are reported in log10.}
    \label{fig:loss_curves}
\end{figure}

Language models are pre-trained using the Huggingface Transformers library \citep{wolf-etal-2020-transformers}.
Hyperparameters are reported in Table \ref{tab:hyperparams}, and loss curves are in Figure \ref{fig:loss_curves}.

\setlength\tabcolsep{3pt}
\begin{table}[ht]
    \centering
    \small
    \renewcommand{\arraystretch}{1.11}
    \begin{tabular}{|>{\raggedright}p{5cm}|r|}
        \hline
        \textbf{Hyperparameter} & \textbf{Value} \\
        \hline
        Layers & 12 \\
        Embedding size & 768 \\
        Hidden size & 768 \\
        Intermediate hidden size & 3072 \\
        Attention heads & 12 \\
        Attention head size & 64 \\
        Activation function & GELU \\
        Vocab size & 50004 \\
        Max sequence length & 128 \\
        Position embedding & Absolute \\
        Batch size & 256 \\
        Train steps & 1M \\
        Learning rate decay & Linear \\
        Warmup steps & 10000 \\
        Learning rate & 1e-4 \\
        Adam $\epsilon$ & 1e-6 \\
        Adam $\beta_1$ & 0.9 \\
        Adam $\beta_2$ & 0.999 \\
        Dropout & 0.1 \\
        Attention dropout & 0.1 \\
        \hline
    \end{tabular}
    \normalsize
    \caption{Language model pre-training hyperparameters \citep{devlin-etal-2019-bert,chang-bergen-2022-word}.}
    \label{tab:hyperparams}
\end{table}

\setlength{\belowcaptionskip}{0.0cm}
\setlength{\belowcaptionskip}{0cm}

Each model takes 2.1 weeks to train on four NVIDIA TITAN Xp GPUs or 2.5 weeks to train on one NVIDIA RTX A6000 GPU.
Including pre-training and inference (evaluation surprisals), our experiments take approximately 2220 hours in A6000 GPU hours.
Computing fitted GAM curves, distances between curves, $n$-gram probabilities, contextual diversities, and POS tags takes approximately 2990 CPU core hours.

\subsection{Checkpoint Strategy}
\label{app:checkpoint-strategy}
Assume that the steps per checkpoint $s(t)$ increases linearly as a function of the current step $t$.
Assume we start with some $s(0) = s_0$ and end with $s(t_1) = s_1$ steps per checkpoint.
Then:\footnote{Assume $t_1 > 0$ and $s_1 > s_0 > 0$.}
\[ s(t) = s_0 + \frac{s_1 - s_0}{t_1} t \]
The rate of checkpoints per step is the inverse of steps per checkpoint, or $1/s(t)$.
Excluding the checkpoint at step zero (i.e. $\text{checkpoints}(0) = 0$), the number of checkpoints at step $t$ is:
\begin{align*}
\text{checkpoints}(t) &= \text{checkpoints}(0) + \int_0^t \frac{1}{s(t)} dt
\end{align*}
\begin{align*}
\hspace{1.5cm} &= \int_0^t \frac{t_1}{s_0 t_1 + (s_1 - s_0) t} dt \\
&= \frac{t_1}{s_1 - s_0} \text{ln} \left( 1 + \frac{s_1 - s_0}{s_0 t_1} t \right)
\end{align*}
By solving for $t$, we can compute the time steps where the number of checkpoints is equal to $n=0,1,2,3, \text{etc}$.
Formally, we can compute the time step $t$ for the $n^{\text{th}}$ checkpoint:
\begin{align*}
n &= \text{checkpoints}(t) \\
n &= \frac{t_1}{s_1 - s_0} \text{ln} \left( 1 + \frac{s_1 - s_0}{s_0 t_1} t \right)
\end{align*}
\begin{align*}
t &= \frac{s_0 t_1}{s_1 - s_0} \left( \bm{e}^{n \left( \frac{s_1 - s_0}{t_1} \right)} - 1 \right) 
\end{align*}
Note that $t$ increases exponentially as a function of the checkpoint number $n$.
For our experiments, we start with $s_0 = s(0) = 100$ steps per checkpoint.
We end with $s_1 = s(1000000) = 25000$ steps per checkpoint at step 1M.
Then, the time step $t$ for the $n^{\text{th}}$ checkpoint is:
\begin{align*}
\text{step}(n) &= \frac{100*1000000}{24900} \left( \bm{e}^{n \left( \frac{24900}{1000000} \right)} - 1 \right)
\end{align*}
We round each $\text{step}(n)$ to the nearest integer, and we save model checkpoints at the selected steps until reaching 1M steps.
Concretely, we save checkpoints at steps:
$\text{step}(1) = 101$,
$\text{step}(2) = 205$, ...,
$\text{step}(221) = 981536$.
We also save one checkpoint at step zero.
In total, we save 222 checkpoints per pre-training run.

\setlength\tabcolsep{3pt}
\begin{table*}[t]
    \centering
    \small
    \renewcommand{\arraystretch}{1.11}

    \begin{tabular}{|>{\raggedright}p{1.1cm}|r|}
        \hline
        \multicolumn{2}{|c|}{Surprisal} \\
        \multicolumn{1}{|l}{\textbf{Tag}} & \textbf{Coef.} \\
        \hline
PART & -1.28 \\
AUX & -1.19 \\
NOUN & -1.17 \\
PUNCT & -1.16 \\
PRON & -1.13 \\
X & -1.12 \\
SYM & -0.89 \\
PROPN & -0.81 \\
ADP & -0.75 \\
VERB & -0.64 \\
NUM & -0.60 \\
SCONJ & -0.53 \\
CCONJ & -0.49 \\
INTJ & -0.47 \\
DET & -0.41 \\
ADJ & -0.35 \\
ADV & -0.11 \\
\hline
L & -0.56 \\
I & -0.26 \\
U & 0.00 \\
B & 0.82 \\
        \hline
    \end{tabular}
\hspace{0.2cm}
        \begin{tabular}{|>{\raggedright}p{1.1cm}|r|}
        \hline
        \multicolumn{2}{|c|}{Var. (steps)} \\
        \multicolumn{1}{|l}{\textbf{Tag}} & \textbf{Coef.} \\
        \hline
INTJ & -0.03 \\
PART & -0.02 \\
X & -0.01 \\
AUX & -0.01 \\
SCONJ & -0.01 \\
NUM & -0.01 \\
VERB & 0.00 \\
PRON & 0.00 \\
ADV & 0.00 \\
DET & 0.00 \\
PROPN & 0.00 \\
PUNCT & 0.00 \\
ADJ & 0.00 \\
ADP & 0.00 \\
CCONJ & 0.01 \\
SYM & 0.01 \\
NOUN & 0.01 \\
\hline
B & -0.01 \\
L & 0.00 \\
U & 0.00 \\
I & 0.02 \\
        \hline
    \end{tabular}
    \hspace{0.2cm}
        \begin{tabular}{|>{\raggedright}p{1.1cm}|r|}
        \hline
        \multicolumn{2}{|c|}{AoA} \\
        \multicolumn{1}{|l}{\textbf{Tag}} & \textbf{Coef.} \\
        \hline
INTJ & -0.30 \\
PUNCT & -0.25 \\
DET & -0.23 \\
NUM & -0.19 \\
X & -0.18 \\
VERB & -0.17 \\
ADJ & -0.17 \\
ADV & -0.15 \\
SYM & -0.15 \\
PROPN & -0.13 \\
PART & -0.12 \\
CCONJ & -0.11 \\
SCONJ & -0.09 \\
NOUN & -0.07 \\
PRON & -0.06 \\
AUX & -0.04 \\
ADP & 0.01 \\
\hline
U & 0.00 \\
I & 0.00 \\
L & 0.03 \\
B & 0.03 \\
        \hline
    \end{tabular}
\hspace{0.2cm}
        \begin{tabular}{|>{\raggedright}p{1.1cm}|r|}
        \hline
        \multicolumn{2}{|c|}{Forgettability} \\
        \multicolumn{1}{|l}{\textbf{Tag}} & \textbf{Coef.} \\
        \hline
INTJ & -0.36 \\
SCONJ & -0.25 \\
ADV & -0.22 \\
VERB & -0.19 \\
NUM & -0.12 \\
PART & -0.11 \\
X & -0.09 \\
ADJ & -0.08 \\
PRON & -0.03 \\
AUX & -0.01 \\
ADP & 0.05 \\
NOUN & 0.09 \\
CCONJ & 0.14 \\
PUNCT & 0.22 \\
PROPN & 0.27 \\
SYM & 0.29 \\
DET & 0.50 \\
\hline
U & 0.00 \\
B & 0.43 \\
L & 0.47 \\
I & 0.67 \\
        \hline
    \end{tabular}
\hspace{0.2cm}
        \begin{tabular}{|>{\raggedright}p{1.1cm}|r|}
        \hline
        \multicolumn{2}{|c|}{Var. (runs)} \\
        \multicolumn{1}{|l}{\textbf{Tag}} & \textbf{Coef.} \\
        \hline
INTJ & -0.23 \\
NUM & -0.15 \\
VERB & -0.08 \\
ADV & -0.05 \\
SCONJ & -0.05 \\
AUX & -0.02 \\
ADJ & -0.02 \\
PART & 0.02 \\
PRON & 0.03 \\
SYM & 0.04 \\
DET & 0.07 \\
X & 0.07 \\
CCONJ & 0.08 \\
PUNCT & 0.11 \\
ADP & 0.12 \\
NOUN & 0.15 \\
PROPN & 0.15 \\
\hline
U & 0.00 \\
B & 0.11 \\
L & 0.30 \\
I & 0.41 \\
        \hline
    \end{tabular}
    
    \normalsize
    \caption{Part-of-speech (POS) tag coefficients when predicting each learning curve metric, after accounting for other predictors (\S\ref{sec:predictors-regressions}). POS tags use the Universal POS tags \citep{nivre-etal-2020-universal}, and we include a feature indicating whether a token is the first token (B), intermediate token (I), last token (L), or only token (U) in a word.}
    \label{tab:pos-coefs}
\end{table*}

\subsection{Part-of-Speech (POS) Coefficients}
\label{app:pos-coefs}
In Table \ref{tab:pos-coefs}, we report coefficients for all POS tags when predicting each learning curve metric, after accounting for other predictors (\S\ref{sec:predictors-regressions}).

\end{document}